\documentclass[letterpaper,journal]{IEEEtran}
\usepackage{amsmath,amsfonts,amssymb}
\usepackage{mathrsfs}
\usepackage{algpseudocode}
\usepackage{algorithm}
\usepackage{float}
\usepackage{array}
\usepackage[caption=false,font=normalsize,labelfont=sf,textfont=sf]{subfig}
\usepackage{textcomp}
\usepackage{stfloats}
\usepackage{cuted}
\usepackage{url}
\usepackage{verbatim}
\usepackage{graphicx}
\usepackage{cite}
\usepackage{color}
\usepackage{booktabs}
\usepackage{multirow}
\usepackage{amsthm}
\newtheoremstyle{mythm}
  {}
  {}
  {\upshape}
  {0pt}
  {\bfseries}
  {.}
  {1em}
  {}
\theoremstyle{mythm}
\newtheorem{theorem}{Theorem}
\newtheorem{assumption}{Assumption}
\newtheorem{corollary}{Corollary}
\newtheorem{lemma}{Lemma}
\newtheorem*{remark}{Remark}

\begin{document}
\bstctlcite{IEEEcontrol}

\title{CARE-VI: Conservative Adaptive Reliability Estimation for Value Improvement in Off-Policy Actor-Critic Learning}

% Author ORCIDs for submission metadata:
% Xiang Zou: 0009-0004-7107-6384
% Shengzhu Shi: 0009-0001-0356-902X
% Junqi Gao: 0009-0007-1644-5812
% Zhichang Guo: 0000-0001-8428-1396
\author{Xiang~Zou, Shengzhu~Shi, Junqi~Gao, and Zhichang~Guo%
\thanks{\textit{(Corresponding author: Zhichang Guo.)}}%
\thanks{The authors are with School of Mathematics, Harbin Institute of Technology, Harbin 150001, China (e-mail: mathgzc@hit.edu.cn).}}

\markboth{}{}

\maketitle

\begin{abstract}
Reliable temporal-difference targets are central to off-policy actor-critic learning. Direct value improvement refines the next-state target with alternative actions, but the reliability of this refinement depends on how candidate actions are ranked, reviewed, and weighted. Noisy rankings may force premature candidate commitment, reusing selection scores may bias target valuation, and fixed enhancement weights may amplify weak evidence.
To address these risks, we develop Conservative Adaptive Ranking and Screening (CARS), which retains an ordered candidate prefix within a preset budget and narrows it only when the observed boundary gap exceeds a disagreement-scaled uncertainty radius. Selector-Evaluator Value Assessment (SEVA) uses selector critics to order candidates and a separately parameterized evaluator critic to review the selected value, then caps the reviewed value at the selector reference. Dynamic Adaptive Risk-aware Enhancement (DARE) then regulates each residual correction using candidate reliability, the gap between selector and evaluator signals, and a finite stage factor.
Together, CARS, SEVA, and DARE form CARE-VI, an evidence-regulated target construction framework that preserves the backbone interfaces for critic regression and actor updates. The analysis bounds the CARS boundary error, the SEVA selected-value overestimation, and the one-sided deviation of the DARE residual displacement from its population counterpart, and establishes fixed-policy recovery after the finite-stage perturbation ends. Experiments with SAC, TD3, and TD7 on four MuJoCo tasks show that CARE-VI achieves the highest mean return in all twelve settings. Grouped ablations and scalar diagnostics support the roles of the three components in improving target reliability.
\end{abstract}


\section{Introduction}
\IEEEPARstart{O}{ff-policy} actor-critic methods have achieved remarkable success across continuous control domains, including robotic manipulation and locomotion control~\cite{sutton1999policy,silver2014deterministic,lillicrap2016ddpg,fujimoto2018td3,haarnoja2018sac,li2024wpvop}. Temporal-difference (TD) targets provide stable supervision to the critic for action value estimation through Bellman updates~\cite{lillicrap2016ddpg,fujimoto2018td3,haarnoja2018sac,kuznetsov2020tqc,chen2021redq,oren2025value}. However, unreliable target values can distort critic estimates and subsequently degrade policy improvement~\cite{tsitsiklis1997analysis,fujimoto2018td3,huang2025dorlvc}.

There are two complementary approaches for improving target reliability. Some methods use direct correction of scalar estimates, whereas others use information beyond a single scalar estimate. Double Q-learning and Maxmin Q-learning reduce estimation bias, whereas target averaging reduces variance~\cite{vanhasselt2010doubleq,vanhasselt2016doubledqn,lan2020maxmin,anschel2017averaged,zhang2017weighted}. Clipped target estimation and weakly pessimistic value estimation control approximation error and optimism, respectively~\cite{fujimoto2018td3,li2024wpvop}. Beyond scalar correction, distributional critics parameterize returns via categorical or quantile distributions~\cite{bellemare2017distributional,rowland2018categorical,dabney2018distributional,dabney2018implicit,yang2019fqf,valencia2025ctd4,bai2024mqn}. Offline reinforcement learning (RL) methods typically rely on support constraints, uncertainty penalties, ensemble mixtures, and data-dependent weighting to address distribution shift~\cite{fujimoto2019bcq,kumar2019bear,kumar2020cql,yu2020mopo,kidambi2020morel,agarwal2020rem,an2021edac,kostrikov2022iql,wu2021uwac,huang2024mpe}. The errors caused by data sparsity and Bellman uncertainty negatively impact value estimates and lead to unreliable decisions~\cite{zhang2026ieeds,jiang2026luc}. Randomized value functions and deep ensembles use randomization and disagreement, whereas uncertainty Bellman equations use recursive updates to propagate uncertainty over time~\cite{osband2016deep,osband2018randomized,lakshminarayanan2017deepensembles,odonoghue2018uncertainty,chen2021redq,lee2021sunrise,an2021edac}. In reality, critics will not exactly converge to truth, and this uncertainty should be taken into account when building targets~\cite{fujimoto2018td3,odonoghue2018uncertainty,an2021edac}.

Recently, direct value improvement has emerged as a promising way to refine the next state target. Representative methods include Value-Improved Actor Critic (VIAC) with a value improvement operator and Blended Exploitation and Exploration (BEE) with successful replay buffer actions~\cite{oren2025value,ji2024seizing}. These methods enlarge or refine the candidate source, but they do not jointly regulate three decisions made inside target construction. A noisy ranking can force premature commitment within the retained-candidate budget, the score that favored an action can be reused as its target value, and a fixed coefficient can inject the resulting residual without regard to its supporting evidence~\cite{vanhasselt2010doubleq,fujimoto2018td3,lee2021sunrise}. Because these decisions precede bootstrapping, their errors can enter subsequent critic and policy updates~\cite{tsitsiklis1997analysis,fujimoto2018td3}. Direct value improvement therefore requires coordinated control over retained width, reviewed value, and residual injection.

When candidate gaps are comparable to critic disagreement, a hard top-one decision commits to distinctions that the available evidence may not resolve~\cite{fujimoto2018td3,an2021edac}. Within a preset candidate budget, retaining a wider leading prefix until an observed boundary gap exceeds its uncertainty scale reduces this premature commitment. Following this observation, we propose Conservative Adaptive Ranking and Screening (CARS), which adapts the retained width with a disagreement-scaled uncertainty radius. Candidate screening alone cannot control the error introduced when a favored score is reused for target valuation~\cite{vanhasselt2010doubleq,vanhasselt2016doubledqn}. A selector should establish the primary order, while a separately parameterized evaluator supplies complementary evidence and reviews the final value. We accordingly develop Selector-Evaluator Value Assessment (SEVA), which combines selector-led ordering with auxiliary evaluator evidence and caps the reviewed value by the selector reference. The remaining residual should also receive less weight when candidate evidence is weak or selector and evaluator signals diverge. Dynamic Adaptive Risk-aware Enhancement (DARE) is developed by applying uncertainty-weighted target estimation~\cite{buckman2018steve,wu2021uwac,lee2021sunrise} to regulate each residual through candidate reliability, the selector-evaluator evidence gap, and a finite training stage.

The three components form Conservative Adaptive Reliability Estimation for Value Improvement (CARE-VI), an evidence-regulated target construction framework that retains existing critic-regression and actor-update interfaces. Its analysis follows the target computation itself. CARS controls observed boundary error, SEVA bounds selected-value overestimation after data-dependent choice, and DARE limits the one-sided deviation of its residual displacement from the population-supported counterpart. The finite CARE-VI perturbation further yields fixed-policy recovery after the active interval.

The main contributions of this article are summarized below.
\begin{enumerate}
\renewcommand{\labelenumi}{\arabic{enumi})}
    \item We propose CARS, SEVA, and DARE to control commitment within a candidate budget, perform selector-led ordering with capped evaluator review, and regulate residual magnitude and duration.

    \item We integrate the three components into CARE-VI, an evidence-regulated framework that retains the backbone critic-regression and actor-update interfaces and admits component-aligned error bounds together with fixed-policy evaluation recovery.

    \item CARE-VI obtains the highest reported mean return in all tested combinations of three backbones and four MuJoCo tasks. Grouped ablations quantify the effect of removing each component, and scalar diagnostics isolate the failure modes that motivate capped review and evidence-regulated residuals.
\end{enumerate}

\section{Background}

In this section, we start with providing preliminaries on off-policy actor-critic learning and direct value improvement for next state target construction.

\subsection{Problem Setting}

Consider a Markov decision process (MDP)
$\mathcal{M}=\langle \mathcal{S},\mathcal{A},\mathcal{P},r,\gamma\rangle$,
where $\mathcal{S}$ is the state space, $\mathcal{A}\subseteq \mathbb{R}^d$ is the continuous action space, $\mathcal{P}$ is the transition kernel, $r$ is the reward function, and $\gamma\in(0,1)$ is the discount factor. The policy $\pi_\theta$ maximizes
\[
J(\pi_\theta)=
\mathbb{E}_{\pi_\theta,\mathcal{P}}
\left[
\sum_{t=0}^{\infty}\gamma^t r(s_t,a_t)
\right],
\]
where $a_t\sim\pi_\theta(\cdot|s_t)$ and $s_{t+1}\sim\mathcal{P}(\cdot|s_t,a_t)$.

As an off-policy actor-critic method, the base algorithm maintains a replay buffer $\mathcal{D}$ and constructs the bootstrapped TD target as~\cite{lillicrap2016ddpg,fujimoto2018td3,haarnoja2018sac}
\[
y_t = r_t+\gamma (1-\mathrm{done}_t) V_{\mathrm{tar}}(s_{t+1}),
\]
where $y_t$ is the bootstrapped TD target used in the critic loss, $V_{\mathrm{tar}}$ denotes the next state target value and $\mathrm{done}_t$ is the termination indicator. The factor $(1-\mathrm{done}_t)$ determines whether the future value term is retained in the TD target. It is kept for nonterminal transitions and removed at terminal transitions. For notational convenience, it is omitted below.

The critic $Q_\phi$ is updated by minimizing
\[
\begin{aligned}
\mathcal{L}_{Q_{\phi}}=
\mathbb{E}_{(s_t,a_t,r_t,s_{t+1})\sim\mathcal{D}}
\left[
\left(Q_{\phi}(s_t,a_t)-y_t\right)^2
\right].
\end{aligned}
\]
The actor is improved using the current critic~\cite{konda1999actor},
\[
\begin{aligned}
\mathcal{J}_{\pi}(\theta)=
\mathbb{E}_{\substack{s_t\sim\mathcal{D}\\
\tilde{a}\sim\pi_\theta(\cdot|s_t)}}
\left[
\Psi_{\phi}(s_t,\tilde{a})
\right],
\end{aligned}
\]
where $\Psi_{\phi}$ abstracts the base policy-improvement term. Target networks use Polyak averaging~\cite{lillicrap2016ddpg,fujimoto2018td3},
\[
\phi' \leftarrow \tau \phi+(1-\tau)\phi',
\]
with $\tau\in(0,1)$. Regularized MDP theory studies modified Bellman and
policy-improvement operators~\cite{geist2019theory}. CARE-VI keeps the actor
objective unchanged and modifies only $V_{\mathrm{tar}}(s_{t+1})$.

\subsection{Target Value Construction}

Without additional value improvement, the base algorithm constructs a
continuation value $B_0(s_{t+1})$ according to its own target
rule~\cite{lillicrap2016ddpg,fujimoto2018td3,haarnoja2018sac}, yielding
\[
y_t=r_t+\gamma B_0(s_{t+1}).
\]
Twin Delayed Deep Deterministic Policy Gradient (TD3) uses clipped double target critics and target policy smoothing, but still
evaluates a base-policy target action~\cite{fujimoto2018td3}.

To construct an enhanced next state target, direct value improvement replaces the base next state value with an enhanced value.
\[
\begin{aligned}
&y_t^{\mathrm{VI}} =
r_t+\gamma V_{\mathrm{VI}}(s_{t+1}),
\\
&V_{\mathrm{VI}}(s_{t+1})
=
I_V\!\left(
s_{t+1},
\pi_{\theta'},
Q_{\phi_1'},
Q_{\phi_2'}
\right),
\end{aligned}
\]
where $I_V$ denotes a value improvement operator that maps the next state, the target policy, and the target critics to an enhanced next state value~\cite{oren2025value,ji2024seizing}. A typical instance samples a candidate set $\mathcal{A}(s_{t+1})$ and returns a critic-based candidate value such as $\max_{a\in\mathcal{A}(s_{t+1})}\min_{\ell\in\{1,2\}}Q_{\phi_\ell'}(s_{t+1},a)$~\cite{oren2025value}. Here $\pi_{\theta'}$ is the target policy, and $Q_{\phi_1'}$ and $Q_{\phi_2'}$ are the target critics used in this next state value. This construction does not by itself regulate candidate commitment under ranking uncertainty~\cite{fujimoto2018td3}, the reuse of favored estimates in target valuation~\cite{vanhasselt2010doubleq}, or the weight assigned to a residual with limited evidence~\cite{lee2021sunrise}. The three components developed below act on these stages of target construction.

\section{Methodology}
\label{sec:methodology}

This section develops evidence-regulated target construction through three
successive components for retained width, value review, and residual
injection.

\subsection{Definitions and Assumptions}
\label{subsec:definitions_assumptions}

We begin with the notation and assumptions used in the analysis. 

Candidate ordering uses the two selector critics, whereas value review uses a
separately parameterized evaluator critic. The selector mean
treats the two critics symmetrically and keeps their common value level
separate from the disagreement signal used below. Throughout this section,
$t$ indexes a target-construction update. The experimental schedule
instantiates this index with the environment-step counter that triggers the
corresponding update. For a target construction call indexed by $t$ and next
state $s'$, the selector reference is defined by
\begin{equation}
\bar Q_t(s',a)
=
\frac{Q'_{1,t}(s',a)+Q'_{2,t}(s',a)}{2}.
\label{eq:analysis_selector_evaluator_values}
\end{equation}
Here, $Q'_{\ell,t}=Q_{\phi_{\ell,t}'}$ for $\ell\in\{1,2\}$ and
$Q'_{\mathrm{eval},t}=Q_{\phi_{\mathrm{eval},t}'}$. For TD7, its two native
critic outputs instantiate $Q'_{1,t}$ and $Q'_{2,t}$, while the twin outputs of
the separately parameterized evaluator are averaged when the evaluator role is
instantiated.

The candidate proposal follows the target-action convention of the backbone.
Soft Actor-Critic (SAC) draws actions from its stochastic policy. TD3 and TD7 instead perturb the
deterministic target action with zero mean Gaussian noise, clip each noise
coordinate at $c_\pi$, and rescale it by the action half-range
$\mathbf s_{\mathcal A}=(\mathbf a_{\mathrm{high}}-\mathbf
a_{\mathrm{low}})/2$. This produces distinct candidates while retaining the target smoothing
used by deterministic backbones. The unprojected proposal is therefore
\begin{equation}
\begin{aligned}
\widetilde a_j
&\sim
\pi_{\theta'}(\cdot\mid s'),
&&\text{for SAC},\\
\xi_j
&\sim
\mathcal N(\mathbf 0,\sigma_\pi^2\mathbf I),
&&\text{for TD3 and TD7},\\
\widetilde a_j
&=
\mu_{\theta'}(s')+
\operatorname{clip}_{[-c_\pi,c_\pi]}(\xi_j)
\odot\mathbf s_{\mathcal A},
&&\text{for TD3 and TD7}.
\end{aligned}
\label{eq:care_candidate_proposal}
\end{equation}
The action projection and the resulting candidate set are
\begin{equation}
\begin{aligned}
a_j&=\operatorname{clip}_{\mathcal A}(\widetilde a_j),\\
\mathcal A_M(s')&=\{a_1,\ldots,a_M\},
\qquad j\in\{1,\ldots,M\}.
\end{aligned}
\label{eq:cars_candidate_generation}
\end{equation}
Critic disagreement provides a candidate dependent uncertainty proxy in
ensemble value estimation~\cite{lakshminarayanan2017deepensembles,lee2021sunrise,an2021edac}.
A scaled disagreement is subtracted from the selector reference to form the
conservative ranking score. For each candidate $a_j$, the selector disagreement
and conservative ranking score are defined for
$\lambda_{\mathrm{div}}>0$ and $\varepsilon_{\mathrm{unc}}>0$ by
\begin{equation}
\begin{aligned}
d_{j,t}(s')
&=
\left|
Q'_{1,t}(s',a_j)-Q'_{2,t}(s',a_j)
\right|,\\
c_{j,t}(s')
&=
\bar Q_t(s',a_j)-\lambda_{\mathrm{div}}d_{j,t}(s').
\end{aligned}
\label{eq:cars_score}
\end{equation}
The additive score keeps the selector reference and the uncertainty penalty
as separate terms, with $\lambda_{\mathrm{div}}$ controlling their tradeoff.
The positive calibrated floor $\varepsilon_{\mathrm{unc}}$ covers common
selector fluctuations that disagreement cannot represent and prevents the
local scale from degenerating when the two selector outputs are numerically
close. All random quantities are defined on a probability space
$(\Omega,\mathcal F,\mathbb P)$, where
$\Omega$ is the sample space, $\mathcal F$ is the event sigma algebra, and
$\mathbb P$ is the probability measure.

Let $\mathcal F_t(s')\subseteq\mathcal F$ be the sigma field generated by the
realized candidate set, the candidate disagreement values, and the independent
backbone reference action used by TD3 and TD7. The conditional reference score
is the conditional mean of the observed score and is defined by
\begin{equation}
\bar c_{j,t}(s')
=
\mathbb E
\left[
c_{j,t}(s')
\mid
\mathcal F_t(s')
\right].
\label{eq:analysis_score_error_definitions}
\end{equation}
Let $a_t^\star(s')$ and $\lambda_t(s')$ denote the selected action and the
enhancement coefficient. We use $a_t^0(s')$ for the action generated by the
unmodified backbone target. It is an independent draw from the target policy
for SAC and an independent target-smoothed action for TD3 and TD7. During the
active interval, define $a_t^{\mathrm{ref}}(s')=a_t^\star(s')$ for SAC and
$a_t^{\mathrm{ref}}(s')=a_t^0(s')$ for TD3 and TD7. The critic component used
as the residual reference and its backbone correction are
\begin{equation}
\begin{aligned}
V_{\mathrm{ref},t}(s')
&=
\min_{\ell\in\{1,2\}}
Q'_{\ell,t}\bigl(s',a_t^{\mathrm{ref}}(s')\bigr),\\
h_t(s')
&=
\begin{cases}
\alpha_t\log\pi_{\theta'}\bigl(a_t^\star(s')\mid s'\bigr),
& \text{for SAC},\\
0,
& \text{for TD3 and TD7}.
\end{cases}
\end{aligned}
\label{eq:care_backbone_target_interface}
\end{equation}
The corresponding reference continuation is
$B_{\mathrm{ref},t}(s')=V_{\mathrm{ref},t}(s')-h_t(s')$. The continuation
generated by the unmodified backbone is
\begin{equation}
\begin{aligned}
B_{0,t}(s')
&=
\min_{\ell\in\{1,2\}}
Q'_{\ell,t}\bigl(s',a_t^0(s')\bigr)\\[-0.2em]
&\quad
-\alpha_t\log\pi_{\theta'}\bigl(a_t^0(s')\mid s'\bigr),
&& \text{for SAC},\\
B_{0,t}(s')
&=V_{\mathrm{ref},t}(s'),
&& \text{for TD3 and TD7}.
\end{aligned}
\label{eq:care_original_backbone_continuation}
\end{equation}
Thus, the deterministic backbones retain clipped double evaluation and target
policy smoothing in the residual reference. SAC retains its entropy correction
at the action associated with each continuation. Here, $\alpha_t$ is the SAC
temperature parameter. To prevent an evaluator estimate above the selector
reference from increasing the target, we upper bound the reviewed next state
value of the selected action by that reference. The resulting capped target is
defined by
\begin{equation}
V_{\mathrm{cap},t}(s')
=
\min\left\{
Q'_{\mathrm{eval},t}\bigl(s',a_t^\star(s')\bigr),
\bar Q_t\bigl(s',a_t^\star(s')\bigr)
\right\}.
\label{eq:seva_capped_value}
\end{equation}
Using the residual reference introduced above, define the residual introduced
by the capped target as
\begin{equation}
R_t(s')
=
V_{\mathrm{cap},t}(s')-V_{\mathrm{ref},t}(s').
\label{eq:analysis_target_residuals}
\end{equation}

Let $\pi_t$ denote the policy represented by the target actor at update $t$.
The corresponding population action value is
\begin{equation}
q_t(s',a)=Q^{\pi_t}(s',a),
\label{eq:analysis_population_action_value}
\end{equation}
where $Q^{\pi_t}$ is entropy regularized for SAC and ordinary discounted for
TD3 and TD7. The population value associated with the residual reference is
\begin{equation}
v_{\mathrm{ref},t}(s')
=q_t\bigl(s',a_t^{\mathrm{ref}}(s')\bigr).
\label{eq:analysis_population_base_value}
\end{equation}
The critic errors used below are
\begin{equation}
\varepsilon_{\ell,t}(s',a)=Q'_{\ell,t}(s',a)-q_t(s',a),
\qquad
\ell\in\{1,2,\mathrm{eval}\}.
\label{eq:analysis_value_error_definitions}
\end{equation}

For the bounded critic space used in the fixed-policy analysis, let
$\mathcal Q_b$ be the Banach space of bounded action value functions equipped
with the supremum norm. A critic triple
$\mathbf Q=(Q_1,Q_2,Q_{\mathrm{eval}})$ belongs to $\mathcal Q_b^3$, whose
product-space norm is
\begin{equation}
\left\|\mathbf Q\right\|_\infty
=
\max_{\ell\in\{1,2,\mathrm{eval}\}}
\left\|Q_\ell\right\|_\infty.
\label{eq:analysis_critic_triple_norm}
\end{equation}

The probability statements below condition on the realized candidate set,
reference action, and disagreement values, while retaining the critic
estimation randomness induced by initialization, replay sampling, and
stochastic optimization.

Learned critics can exhibit conditional bias caused by finite data and function
approximation, together with random fluctuations from stochastic training~\cite{fujimoto2018td3}. We therefore use one biased conditional moment
generating function bound to control both components without assuming the tail
guarantee derived later. This condition holds whenever the conditional critic
error is the sum of a bias with bounded magnitude and a centered fluctuation
that is conditionally bounded or Gaussian, since either fluctuation admits a
finite sub-Gaussian moment envelope.

\begin{assumption}[Calibrated biased sub-Gaussian critic error]
\label{ass:critic_error_envelope}
Fix $t$ and $s'$. For every $\ell\in\{1,2,\mathrm{eval}\}$ and every
$a\in\mathcal A_M(s')\cup\{a_t^{\mathrm{ref}}(s')\}$, let
$b_{\ell,t}(s',a)\geq0$ and $\nu_{\ell,t}(s',a)>0$ be finite and
$\mathcal F_t(s')$-measurable. For every $\theta\in\mathbb R$,
\begin{equation}
\begin{aligned}
&
\mathbb E\!\left[
\exp\!\left(\theta\varepsilon_{\ell,t}(s',a)\right)
\middle|\mathcal F_t(s')
\right]\\
&\qquad\leq
\exp\!\left(
|\theta|b_{\ell,t}(s',a)
+\frac{\theta^2\nu_{\ell,t}^2(s',a)}{2}
\right).
\end{aligned}
\label{eq:analysis_critic_error_envelope}
\end{equation}
\end{assumption}

Ensemble disagreement is widely used as a proxy for value uncertainty
~\cite{lakshminarayanan2017deepensembles,lee2021sunrise,an2021edac}.
It nonetheless measures differential error and cannot represent a fluctuation
shared by both selector critics. CARS therefore uses candidatewise calibration
before sorting and obtains simultaneous control through a union bound, which
does not require errors at different candidates to be independent. In the
controlled diagnostic, independent Gaussian selector noise with standard
deviation $0.05$ gives the exact conditional mean-noise scale
$0.05/\sqrt{2}$, which is conservatively covered by the experimental floor
$0.05$. An independent confirmation passed all 12 tail checks and all 260
moment checks in both the baseline and common-mode Gaussian settings,
including the near-zero-disagreement region, supporting the calibration within
the stated controlled scope.

\begin{assumption}[Calibrated candidate-score fluctuation]
\label{ass:score_error_calibration}
Fix $t$ and $s'$. Let $\varepsilon_{\mathrm{unc}}>0$ be an uncertainty floor
fixed before evaluation and checked on independent held-out data. For every $j\in\{1,\ldots,M\}$ and
every $\theta\in\mathbb R$,
\begin{equation}
\begin{aligned}
&
\mathbb E\!\left[
\exp\!\left(
\theta[c_{j,t}(s')-\bar c_{j,t}(s')]
\right)
\middle|\mathcal F_t(s')
\right]\\
&\qquad\leq
\exp\!\left[
\frac{\theta^2}{2}
\left(
\frac{d_{j,t}^2(s')}{4}+\varepsilon_{\mathrm{unc}}^2
\right)
\right].
\end{aligned}
\label{eq:analysis_score_error_calibration}
\end{equation}
\end{assumption}

\subsection{Conservative Adaptive Ranking and Screening (CARS)}
\label{subsec:cars}

Ranking candidates with noisy selector scores can reverse a boundary order and
force premature commitment within the retained-support budget. Under the local
noise model, uncertainty in an observed boundary gap depends on the
disagreement scales of the leading candidate and the boundary candidate. The
retained prefix should therefore be narrowed within that budget only when the observed separation
exceeds an uncertainty radius that covers both scales. Motivated by this
relation, we develop Conservative Adaptive Ranking and Screening (CARS).
CARS penalizes the selector reference with critic disagreement and ranks the
resulting scores. It then selects a retained width between $K_{\min}$ and
$K_{\max}$ according to the observed boundary evidence.

In practice, CARS uses \eqref{eq:care_candidate_proposal} and
\eqref{eq:cars_candidate_generation} to draw $M$ actions from the
backbone-specific proposal and form the candidate set. It then uses
\eqref{eq:cars_score} to compute the conservative ranking score and selector
disagreement scale of each candidate. The parameter $\lambda_{\mathrm{div}}$
controls the disagreement penalty. For a fixed selector reference, the higher
the selector disagreement, the lower the resulting ranking score.

The candidates are sorted in descending order according to $c_{j,t}(s')$.
The corresponding actions, scores, reference scores, and disagreements are
denoted by $a_{(j)}$, $c_{(j),t}(s')$, $\bar c_{(j),t}(s')$, and
$d_{(j),t}(s')$. The ordered scores satisfy
\begin{equation}
c_{(1),t}(s')
\geq
c_{(2),t}(s')
\geq
\cdots
\geq
c_{(M),t}(s').
\label{eq:cars_score_order}
\end{equation}
Here, $\bar c_{(j),t}(s')$ is the conditional reference score attached to the
candidate at observed rank $j$.

To determine whether a prefix boundary is supported, CARS searches a set of
admissible retained widths. For each width, the observed boundary gap compares
the leading candidate with the first excluded candidate. The boundary scale
uses the largest selector disagreement among the leading candidate and the
examined prefix. Given integers $1\leq K_{\min}\leq K_{\max}<M$ and a failure
probability $\delta\in(0,1)$, define
\begin{equation}
\begin{aligned}
\mathcal K
&=
\{K_{\min},\ldots,K_{\max}\},\\
g_{k,t}(s')
&=
c_{(1),t}(s')-c_{(k+1),t}(s'),\\
u_{k,t}(s')
&=
\max_{1\leq j\leq k+1}
\sqrt{
\frac{d_{(1),t}^{2}(s')+d_{(j),t}^{2}(s')}{4}
    +2\varepsilon_{\mathrm{unc}}^2
},\\
z_\delta
&=
\sqrt{
2\log\frac{2M}{\delta}
}.
\end{aligned}
\label{eq:cars_boundary_quantities}
\end{equation}
A width is certified when its observed boundary gap exceeds the disagreement-scaled
radius $\sqrt{2}z_\delta u_{k,t}(s')$. CARS collects all certified widths in the
stopping set
\begin{equation}
\mathcal S_{\mathrm{stop},t}(s')
=
\left\{
k\in\mathcal K
\mathrel{}
\middle|
\mathrel{}
g_{k,t}(s')
>
\sqrt{2}z_\delta u_{k,t}(s')
\right\}.
\label{eq:cars_stop_set}
\end{equation}
To obtain the smallest prefix supported by the observed boundary, CARS chooses
the first certified width. If no width is certified, it uses $K_{\max}$. The
retained width is therefore determined by
\begin{equation}
K_t(s')
=
\begin{cases}
\min \mathcal S_{\mathrm{stop},t}(s'),
&
\mathcal S_{\mathrm{stop},t}(s')\neq\varnothing,\\
K_{\max},
&
\mathcal S_{\mathrm{stop},t}(s')=\varnothing.
\end{cases}
\label{eq:cars_k_rule}
\end{equation}
This width determines the candidates passed to SEVA. Given $K_t(s')$, the
retained candidate set is
\begin{equation}
\mathcal A_{K_t}(s')
=
\{a_{(1)},\ldots,a_{(K_t(s'))}\}.
\label{eq:cars_retained_set}
\end{equation}
When the stopping set is empty, $K_t(s')=K_{\max}$ and the largest allowed
prefix is passed to SEVA. Thus, CARS adapts commitment within the prescribed
top-$K_{\max}$ support budget.

For each admissible width, let
\begin{equation}
D_{k,t}(s')
=
\bar c_{(1),t}(s')-\bar c_{(k+1),t}(s')
\label{eq:cars_reference_gap}
\end{equation}
denote the conditional reference gap at the observed boundary. The joint
boundary event is defined as
\begin{equation}
\mathcal E_{C,t}(s')
=
\bigcap_{k\in\mathcal K}
\left\{
\left|g_{k,t}(s')-D_{k,t}(s')\right|
\leq
\sqrt{2}z_\delta u_{k,t}(s')
\right\}.
\label{eq:cars_boundary_event}
\end{equation}
The observed boundary error is controlled by the following result.
\begin{theorem}[CARS observed boundary error bound]
\label{thm:cars_retention}
Under Assumption~\ref{ass:score_error_calibration}, for every $\delta\in(0,1)$ and
conditional on $\mathcal F_t(s')$,
\begin{equation}
\mathbb P\left(
\mathcal E_{C,t}(s')
\mid
\mathcal F_t(s')
\right)
\geq
1-\delta.
\label{eq:cars_boundary_gap_transfer}
\end{equation}
\end{theorem}

\begin{remark}
Theorem~\ref{thm:cars_retention} controls the deviation between every observed
boundary gap and its conditional reference gap. The radius grows with the
selector disagreement, so the stopping rule demands a larger observed separation
at a less certain boundary.
\end{remark}

\begin{corollary}[Sufficient condition for CARS stopping]
\label{cor:cars_stopping_condition}
On the event $\mathcal E_{C,t}(s')$, if an admissible width $k$ satisfies
\begin{equation}
D_{k,t}(s')
>
2\sqrt{2}z_\delta u_{k,t}(s'),
\label{eq:cars_separation_condition}
\end{equation}
then $k\in\mathcal S_{\mathrm{stop},t}(s')$ and $K_t(s')\leq k$.
Conversely, every $k\in\mathcal S_{\mathrm{stop},t}(s')$ satisfies
$D_{k,t}(s')>0$.
\end{corollary}

\subsection{Selector-Evaluator Value Assessment (SEVA)}
\label{subsec:seva}

Reusing a selected critic score directly as the target value couples
bootstrapping to the evidence that favored the action in the first
place~\cite{vanhasselt2010doubleq,vanhasselt2016doubledqn}. The selector should
therefore establish the candidate order, while a separately parameterized
evaluator supplies complementary evidence and reviews the selected value. To
realize these differentiated roles, we develop Selector-Evaluator Value
Assessment (SEVA). SEVA receives the retained prefix in
\eqref{eq:cars_retained_set}, preserves the selector-led order, and uses the
evaluator for auxiliary selection evidence and capped value review.

The auxiliary review evaluates the complete ordered candidate pool. This
provides a common normalization reference before CARS applies the retained
width. The selector scores and evaluator values may have different scales. We use
standardized scores for the two quantities before fusion. Computing their
statistics over the full pool makes the reference independent of the retained
width. A positive constant $\varepsilon_{\mathrm{std}}$ stabilizes both scales
when the candidate values have little variation. The selector mean, evaluator
mean, selector scale, and evaluator scale are defined, respectively, by
\begin{equation}
\begin{aligned}
\mu_t^c(s')
&=
\frac{1}{M}
\sum_{j=1}^{M}
c_{(j),t}(s'),\\
\mu_t^e(s')
&=
\frac{1}{M}
\sum_{j=1}^{M}
Q'_{\mathrm{eval},t}(s',a_{(j)}),\\
s_t^c(s')
&=
\sqrt{
\frac{1}{M}
\sum_{j=1}^{M}
\left(
c_{(j),t}(s')-\mu_t^c(s')
\right)^2
+
\varepsilon_{\mathrm{std}}^2
},\\
s_t^e(s')
&=
\sqrt{
\frac{1}{M}
\sum_{j=1}^{M}
\left(
Q'_{\mathrm{eval},t}(s',a_{(j)})-\mu_t^e(s')
\right)^2
+
\varepsilon_{\mathrm{std}}^2
}.
\end{aligned}
\label{eq:seva_standardization_statistics}
\end{equation}
The standardized selector score and standardized evaluator score are defined,
respectively, by
\begin{equation}
\begin{aligned}
\widetilde c_{j,t}^{\mathrm{sel}}(s')
&=
\frac{
c_{(j),t}(s')-\mu_t^c(s')
}{
s_t^c(s')
},\\
\widetilde e_{j,t}^{\mathrm{eval}}(s')
&=
\frac{
Q'_{\mathrm{eval},t}(s',a_{(j)})-\mu_t^e(s')
}{
s_t^e(s')
},
\qquad
j\in\{1,\ldots,M\}.
\end{aligned}
\label{eq:seva_standardized_scores}
\end{equation}

After standardization, a convex combination places selection and review
evidence on the same scale. This form retains both evidence sources and reduces
to a score based only on selector evidence or only on evaluator evidence at
$w=1$ or $w=0$, respectively. Let $w\in[0,1]$ control their tradeoff. SEVA
calculates the review score for the retained candidates as
\begin{equation}
\begin{aligned}
f_{j,t}(s')
&=
w\widetilde c_{j,t}^{\mathrm{sel}}(s')
+
(1-w)\widetilde e_{j,t}^{\mathrm{eval}}(s'),\\
w&\in[0,1],
\qquad
j\in\{1,\ldots,K_t(s')\}.
\end{aligned}
\label{eq:seva_fusion_score}
\end{equation}
SEVA selects a candidate with the largest review score. The minimum over the
maximizing indices provides a fixed rule for equal scores. The selected index
and selected action are determined, respectively, by
\begin{equation}
\begin{aligned}
j_t^\star(s')
&=\min\Bigl\{j\in\{1,\ldots,K_t(s')\}\\
&\qquad\mathrel{\Big|}\ f_{j,t}(s')
=\max_{1\leq \ell\leq K_t(s')}f_{\ell,t}(s')\Bigr\},\\
a_t^\star(s')&=a_{(j_t^\star(s'))}.
\end{aligned}
\label{eq:seva_selection_rule}
\end{equation}
DARE requires the evidence gap and the reviewed residual to refer to the same
candidate. Since both signals use the same full-pool normalization reference,
their absolute difference remains defined for every retained width. We therefore
evaluate the gap at the action selected by SEVA
\begin{equation}
\Delta_t(s')
=
\left|
\widetilde c_{j_t^\star(s'),t}^{\mathrm{sel}}(s')
-
\widetilde e_{j_t^\star(s'),t}^{\mathrm{eval}}(s')
\right|.
\label{eq:seva_selector_evaluator_gap}
\end{equation}
SEVA caps the reviewed value of the selected action according to
\eqref{eq:seva_capped_value}. Substituting this definition into
\eqref{eq:analysis_target_residuals} gives
\begin{equation}
\begin{aligned}
R_t(s')
=
&\bar Q_t\bigl(s',a_t^\star(s')\bigr)
-
V_{\mathrm{ref},t}(s')\\
&+
\min\Bigl\{
Q'_{\mathrm{eval},t}\bigl(s',a_t^\star(s')\bigr)
-
\bar Q_t\bigl(s',a_t^\star(s')\bigr),
0
\Bigr\}.
\end{aligned}
\label{eq:seva_capped_residual_identity}
\end{equation}
The triangle inequality then gives
\begin{equation}
\begin{aligned}
\left|R_t(s')\right|
\leq
&\left|
\bar Q_t\bigl(s',a_t^\star(s')\bigr)-V_{\mathrm{ref},t}(s')
\right|\\
&+
\left|
Q'_{\mathrm{eval},t}\bigl(s',a_t^\star(s')\bigr)
-
\bar Q_t\bigl(s',a_t^\star(s')\bigr)
\right|.
\end{aligned}
\label{eq:seva_absolute_residual_bound}
\end{equation}
The first term measures the displacement of the selector reference from the
reference value, while the second term measures the selector and evaluator gap for
the selected action. This bound specifies the residual magnitude subsequently
regulated by DARE.

To control all candidates before the data-dependent selection, let
$\delta_v\in(0,1)$ and
$z_v=\sqrt{2\log(6(M+1)/\delta_v)}$. The selector and evaluator radii are
defined by
\begin{equation}
\begin{aligned}
\operatorname{rad}^{\mathrm{sel}}_{t}(s')
&=
\max_{1\leq j\leq M}
\frac12\sum_{\ell=1}^{2}
\left[
b_{\ell,t}(s',a_j)+z_v\nu_{\ell,t}(s',a_j)
\right],\\
\operatorname{rad}^{\mathrm{eval}}_{t}(s')
&=
\max_{1\leq j\leq M}
\left[
b_{\mathrm{eval},t}(s',a_j)+z_v\nu_{\mathrm{eval},t}(s',a_j)
\right].
\end{aligned}
\label{eq:analysis_value_error_radii}
\end{equation}
The corresponding joint event controls every critic error over the complete
candidate set and the reference action before either data-dependent selection is
made.
\begin{equation}
\begin{aligned}
&\mathcal E_{v,t}(s')=\\
&\quad\bigcap_{\substack{\ell\in\{1,2,\mathrm{eval}\}\\
a\in\mathcal A_M(s')\cup\{a_t^{\mathrm{ref}}(s')\}}}
\left\{\begin{aligned}
&|\varepsilon_{\ell,t}(s',a)|\\
&\quad\leq b_{\ell,t}(s',a)+z_v\nu_{\ell,t}(s',a)
\end{aligned}\right\}.
\end{aligned}
\label{eq:seva_joint_value_error_event}
\end{equation}
The resulting event is uniform over the realized candidate set and therefore
remains valid for the action selected by CARS and SEVA.
\begin{theorem}[SEVA selected-value overestimation bound]
\label{thm:seva_adaptive_review}
Under Assumption~\ref{ass:critic_error_envelope}, conditional on
$\mathcal F_t(s')$,
$\mathbb P(\mathcal E_{v,t}(s')\mid\mathcal F_t(s'))\geq1-\delta_v$.
On this event, the action selected by SEVA satisfies
\begin{equation}
\begin{aligned}
&V_{\mathrm{cap},t}(s')
-
q_t\bigl(s',a_t^\star(s')\bigr)\\
&\qquad\leq
\min\left\{
\operatorname{rad}^{\mathrm{sel}}_{t}(s'),
\operatorname{rad}^{\mathrm{eval}}_{t}(s')
\right\}.
\end{aligned}
\label{eq:seva_adaptive_review_bound}
\end{equation}
\end{theorem}

\begin{remark}
The event $\mathcal E_{v,t}(s')$ covers the original candidate set before
CARS and SEVA make their data-dependent choices. The cap then transfers the
smaller of the selector and evaluator radii to the positive error of the
selected value.
\end{remark}

\subsection{Dynamic Adaptive Risk-aware Enhancement (DARE)}
\label{subsec:dare}

After candidate screening and capped value review, the residual contribution
still depends on the reliability of the retained boundary and on the agreement
between the two value sources. A fixed coefficient cannot express these local
differences or specify when the correction should be active. DARE therefore
uses the CARS uncertainty, the SEVA evidence gap, and a finite training-stage
factor to regulate the residual entering the target.

For the retained width $K_t(s')$, define the candidate reliability by
\begin{equation}
\begin{aligned}
\zeta_t(s')
&=
\zeta_{\min}+(1-\zeta_{\min})
\exp\left[-\beta_u u_{K_t(s'),t}(s')\right],
\\
 &\qquad \zeta_{\min}\in[0,1),\qquad \beta_u>0.
\end{aligned}
\label{eq:dare_candidate_reliability}
\end{equation}
The reliability decreases smoothly with the uncertainty scale and remains in
$[\zeta_{\min},1]$.

Let $\operatorname{clip}_{[0,1]}(x)=\min\{1,\max\{0,x\}\}$ and define the
normalized training progress by
\begin{equation}
p_t=\operatorname{clip}_{[0,1]}\left(
\frac{t-t_{\mathrm{start}}}{t_{\mathrm{end}}-t_{\mathrm{start}}}
\right).
\label{eq:dare_training_stage}
\end{equation}
Here, $t_{\mathrm{start}},t_{\mathrm{end}}\in\mathbb N_0$ and
$t_{\mathrm{end}}-t_{\mathrm{start}}\geq2$.
The enhancement coefficient is
\begin{equation}
\begin{aligned}
\lambda_t(s')={}&
4\lambda_{\max}p_t(1-p_t)\\
&\times\max\Bigl\{\omega_{\min},
\exp\bigl[
-\beta_\Delta\max\{0,\Delta_t(s')-\Delta_0\}\\
&\hspace{25mm}
-\beta_\zeta\max\{0,\zeta_0-\zeta_t(s')\}
\bigr]\Bigr\},
\end{aligned}
\label{eq:dare-enhancement-coef}
\end{equation}
where $\lambda_{\max}\in(0,1]$, $\Delta_0\geq0$,
$\zeta_0\in(\zeta_{\min},1]$, $\beta_\Delta,\beta_\zeta>0$, and
$\omega_{\min}\in(0,1]$. It follows that
$0\leq\lambda_t(s')\leq\lambda_{\max}$ and
$\lambda_t(s')=0$ for $t\leq t_{\mathrm{start}}$ or
$t\geq t_{\mathrm{end}}$. DARE produces the target displacement
\begin{equation}
\rho_t(s')=\lambda_t(s')R_t(s').
\label{eq:dare-finite-support}
\end{equation}
The corresponding mixed next-state value is
\begin{equation}
\begin{aligned}
V_{\mathrm{mix},t}(s')
&=V_{\mathrm{ref},t}(s')+\rho_t(s')\\
&=(1-\lambda_t(s'))V_{\mathrm{ref},t}(s')\\
&\quad+\lambda_t(s')V_{\mathrm{cap},t}(s').
\end{aligned}
\label{eq:care_mixed_target}
\end{equation}

To compare the DARE displacement with its population counterpart, define
\begin{equation}
\begin{aligned}
G_t(s')
&=q_t\bigl(s',a_t^\star(s')\bigr)-v_{\mathrm{ref},t}(s'),\\
\operatorname{rad}^G_t(s')
&=\max_{\ell\in\{1,2\}}
\left[b_{\ell,t}(s',a_t^{\mathrm{ref}}(s'))
+z_v\nu_{\ell,t}(s',a_t^{\mathrm{ref}}(s'))\right]\\
&\quad+\min\left\{\operatorname{rad}^{\mathrm{sel}}_t(s'),
\operatorname{rad}^{\mathrm{eval}}_t(s')\right\}.
\end{aligned}
\label{eq:dare_population_gain}
\end{equation}
On the event $\mathcal E_{v,t}(s')$, the DARE displacement satisfies
\begin{corollary}[DARE target displacement error bound]
\label{cor:dare_displacement_error}
\begin{equation}
\gamma\left[\rho_t(s')-\lambda_t(s')G_t(s')\right]
\leq\gamma\lambda_t(s')\operatorname{rad}^G_t(s').
\label{eq:dare_displacement_error_bound}
\end{equation}
\end{corollary}

\begin{remark}
The right-hand side is the SEVA error radius multiplied by the same
nonnegative coefficient used to scale the residual. Thus the gate and the
finite stage factor regulate the one-sided target displacement error.
\end{remark}

\subsection{CARE-VI Framework}
\label{subsec:care_vi_framework}

The three component results form a target-construction chain. CARS controls
the evidence required to reduce the retained prefix, SEVA bounds the selected
value after capped review, and DARE transfers the resulting bound to a finite
and evidence-regulated residual. Their joint conditional event satisfies
\begin{equation}
\mathbb P\left(\mathcal E_{C,t}(s')\cap\mathcal E_{v,t}(s')
\mid\mathcal F_t(s')\right)\geq1-\delta-\delta_v.
\label{eq:care_joint_component_event}
\end{equation}

For the fixed-policy analysis, let $\kappa_\pi(\cdot\mid s')$ denote the
time-invariant backbone action-generation kernel. It is the fixed policy
distribution for SAC and the fixed target-smoothing distribution centered at
the deterministic policy action for TD3 and TD7. The SAC temperature is also
held fixed. Let $\mathcal B_0^\pi[\mathbf Q](s')$ denote the expectation of
the unmodified backbone continuation over $\kappa_\pi(\cdot\mid s')$.
We assume that the reward is uniformly bounded and, for SAC, that the expected
entropy correction is uniformly bounded. These conditions make the exact base
evaluation operator below a self-map on $\mathcal Q_b^3$. It is defined
componentwise by
\begin{equation}
\begin{aligned}
(\mathfrak T_0^\pi\mathbf Q)_\ell(s,a)
&=r(s,a)\\
&\quad+\gamma\mathbb E_{s'\sim\mathcal P(\cdot\mid s,a)}
\left[\mathcal B_0^\pi[\mathbf Q](s')\right],\\
&\qquad\ell\in\{1,2,\mathrm{eval}\}.
\end{aligned}
\label{eq:analysis_base_operator}
\end{equation}
For the minimum and average aggregations used by the deterministic and
entropy-regularized backbones, the expected continuation is nonexpansive in the
product maximum norm. Consequently, under the fixed-policy conditions above,
\begin{lemma}[Base fixed-policy evaluation contraction]
\label{lem:base_evaluation_contraction}
\begin{equation}
\left\|\mathfrak T_0^\pi\mathbf Q-
\mathfrak T_0^\pi\widetilde{\mathbf Q}\right\|_\infty
\leq\gamma\left\|\mathbf Q-\widetilde{\mathbf Q}\right\|_\infty.
\label{eq:analysis_base_operator_contraction}
\end{equation}
\end{lemma}
\begin{remark}
Minimum and average value aggregation do not enlarge maximum-norm differences,
and the discount factor reduces the remaining difference by $\gamma$. This is
the standard contraction underlying discounted fixed-policy evaluation
analysis~\cite{tsitsiklis1997analysis,geist2019theory}.
\end{remark}
The lemma gives a unique fixed point, denoted by $\mathbf Q_{\mathrm{base}}^\pi$.

Let $B_{0,t}(s')=\mathcal B_0^\pi[\mathbf Q_t](s')$ and let
$B_{\mathrm{ref},t}(s')=V_{\mathrm{ref},t}(s')-h_t(s')$ be the continuation
associated with the reference action. The complete CARE-VI perturbation is
\begin{equation}
\psi_t(s')=
\begin{cases}
B_{\mathrm{ref},t}(s')-B_{0,t}(s')+\rho_t(s'),
&t_{\mathrm{start}}<t<t_{\mathrm{end}},\\
0,&\text{otherwise}.
\end{cases}
\label{eq:care_total_target_perturbation}
\end{equation}
Therefore, during the active interval,
\begin{equation}
B_{0,t}(s')+\psi_t(s')=B_{\mathrm{ref},t}(s')+\rho_t(s').
\label{eq:care_perturbation_identity}
\end{equation}
For TD3 and TD7, $B_{\mathrm{ref},t}=B_{0,t}$. For SAC, the first term in
$\psi_t$ records the change from the original soft continuation to the selected
reference continuation. Define the transition operator
\begin{equation}
(\mathsf P\psi_t)(s,a)=
\mathbb E_{s'\sim\mathcal P(\cdot\mid s,a)}[\psi_t(s')].
\label{eq:care_transition_residual}
\end{equation}
With $\mathbf1_3=(1,1,1)^\top$, the fitting residual is the scalar
\begin{equation}
\varepsilon_t^{\mathrm{fit}}=
\left\|\mathbf Q_{t+1}-\mathfrak T_0^\pi\mathbf Q_t
-\gamma\mathbf1_3\mathsf P\psi_t\right\|_\infty.
\label{eq:care_fitting_residual}
\end{equation}
The CARE-VI continuation is
\begin{equation}
B_{\mathrm{CARE},t}(s')=B_{0,t}(s')+\psi_t(s').
\label{eq:care_complete_continuation}
\end{equation}
The resulting temporal-difference target is
\begin{equation}
y_t(s,a,r,s')=r+\gamma B_{\mathrm{CARE},t}(s').
\label{eq:care_td_target}
\end{equation}

The critic loss, target update, and actor update retain the backbone interfaces.
For $\ell\in\{1,2,\mathrm{eval}\}$,
\begin{equation}
\mathcal L_{\ell,t}(\phi_{\ell,t})=
\mathbb E_{(s,a,r,s')\sim\mathcal D}
\left[(Q_{\phi_{\ell,t}}(s,a)-y_t(s,a,r,s'))^2\right].
\label{eq:care_critic_loss}
\end{equation}
The target critic update is
\begin{equation}
\phi_{\ell,t+1}'=\tau\phi_{\ell,t+1}+(1-\tau)\phi_{\ell,t}',
\qquad \tau\in(0,1).
\label{eq:care_target_critic_update}
\end{equation}
The actor update uses the fitted critic $\widehat Q_{t+1}$ supplied by the
backbone:
\begin{equation}
\pi_{t+1}=\operatorname{Improve}_{\mathrm{base}}
\left(\pi_t,\widehat Q_{t+1}\right).
\label{eq:care_actor_update}
\end{equation}

The target-construction cost is
\begin{equation}
\mathcal O\left(2M C_{\mathrm{sel}}+M C_{\mathrm{eval}}+M\log M\right),
\label{eq:care_target_complexity}
\end{equation}
with working storage $\mathcal O(M)$.

For the fixed-policy analysis, set
\begin{equation}
\varepsilon_t=\|\psi_t\|_\infty.
\label{eq:care_displacement_magnitude}
\end{equation}
The finite support in \eqref{eq:care_total_target_perturbation} gives
$\psi_t=0$ for $t\geq t_{\mathrm{end}}$.

\begin{theorem}[CARE-VI fixed-policy evaluation recovery]
\label{thm:care_fixed_policy_recovery}
Under the fixed-policy conditions above, consider a critic sequence
$\{\mathbf Q_t\}_{t\geq0}\subset\mathcal Q_b^3$. The bounded continuation
terms and $0\leq\lambda_t(s')\leq\lambda_{\max}$ give
$\max_{0\leq t<t_{\mathrm{end}}}\|\psi_t\|_\infty<\infty$. The finite DARE
schedule gives $\psi_t=0$ for every $t\geq t_{\mathrm{end}}$. Once this schedule
has ended, the target is the base fixed-policy target, and the corresponding
fitting residual satisfies $\lim_{t\to\infty}\varepsilon_t^{\mathrm{fit}}=0$. Then
\begin{equation}
\lim_{T\to\infty}\|\mathbf Q_T-\mathbf Q_{\mathrm{base}}^\pi\|_\infty=0.
\label{eq:care_base_fixed_point_convergence}
\end{equation}
\end{theorem}

\begin{remark}
The contraction removes the finite collection of CARE-VI perturbations after
the active interval, while the fitting residual vanishes as the base target is
tracked. The critic sequence therefore approaches the fixed point of the base
fixed-policy evaluation operator.
\end{remark}

\section{Experiments}

We evaluate CARE-VI on MuJoCo continuous control tasks around three questions.
The main comparison examines whether evidence-regulated target construction
remains effective across different off-policy actor-critic backbones. The
ablation study measures the effect of replacing each grouped component. Two
principle-level scalar diagnostics then isolate selection-score reuse and
evidence-dependent residual scaling. Sensitivity analysis and the DARE schedule
complete the evaluation by examining representative control parameters and the
finite support of the correction.

\subsection{Experimental Setup and Baselines}

Following standard continuous control evaluation, we use four MuJoCo
tasks~\cite{todorov2012mujoco,duan2016benchmarking}.
Table~\ref{tab:mujoco_envs} summarizes their observation dimensions, action
dimensions, and bounded action ranges. The tasks vary in state and action
dimensions while sharing the continuous action interface required by all
compared methods.

\begin{table}[!h]
\centering
\caption{MuJoCo tasks used for evaluation with their observation dimensions,
action dimensions, and elementwise action ranges.}
\label{tab:mujoco_envs}
\vspace{-2mm}
\small
\renewcommand{\arraystretch}{1.15}
\resizebox{\columnwidth}{!}{%
\begin{tabular}{@{}lccc@{}}
\toprule
\textbf{Environment} & \textbf{Observation Dim} & \textbf{Action Dim} & \textbf{Action Range} \\
\midrule
HalfCheetah-v4 & 17 & 6 & $[-1, 1]$ \\
Hopper-v4      & 11 & 3 & $[-1, 1]$ \\
Walker2d-v4    & 17 & 6 & $[-1, 1]$ \\
Ant-v4         & 27 & 8 & $[-1, 1]$ \\
\bottomrule
\vspace{-5mm}
\end{tabular}
}
\end{table}

We instantiate CARE-VI with SAC~\cite{haarnoja2018sac},
TD3~\cite{fujimoto2018td3}, and TD7~\cite{fujimoto2023sale}. Vanilla denotes
the corresponding backbone without direct value improvement, while
VIAC~\cite{oren2025value} and BEE~\cite{ji2024seizing} provide direct value
improvement baselines. Using the same target construction with three backbones
makes it possible to distinguish a method-level pattern from an isolated gain
under one learning algorithm.

For a controlled comparison, all methods share the interaction budget,
evaluation interval, deterministic evaluation policy, and backbone settings.
Each configuration uses $3\times10^6$ environment steps, is repeated with
three random seeds, and is evaluated for ten episodes every $5{,}000$ steps.
Each run is summarized by the mean return over its last ten evaluation
checkpoints, and Table~\ref{tab:performance_comparison} reports the mean and
standard deviation across runs.

\subsection{Performance on MuJoCo}

\begin{figure*}[!t]
    \centering
    \includegraphics[width=\textwidth]{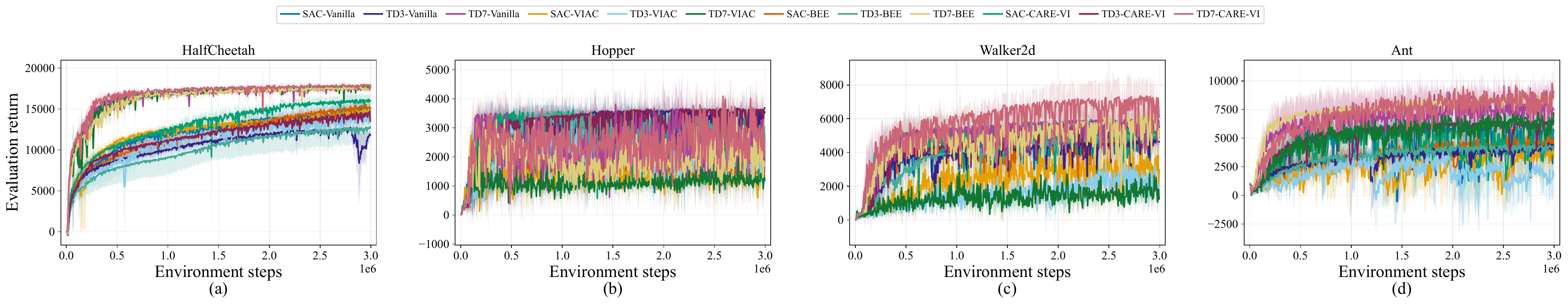}
    \caption{Learning curves on four MuJoCo tasks. Solid curves denote the
    mean evaluation return, and shaded regions indicate one standard deviation
    across three random seeds.}
    \label{fig:performance_curves}
\end{figure*}

Figure~\ref{fig:performance_curves} presents the learning curves for all
backbones, methods, and tasks. The corresponding returns over the last ten
evaluation checkpoints are reported in
Table~\ref{tab:performance_comparison}.

\begin{table*}[!ht]
\centering
\caption{Final performance on MuJoCo. Entries give the mean and standard
deviation across three random seeds after each run is summarized over its last
ten evaluation checkpoints.}
\label{tab:performance_comparison}
\vspace{-2mm}
\small
\begin{tabular}{llcccc}
\toprule
\textbf{Algorithm} & \textbf{Method} & \textbf{Hopper-v4} & \textbf{Ant-v4} & \textbf{HalfCheetah-v4} & \textbf{Walker2d-v4} \\
\midrule
\multirow{4}{*}{SAC} & Vanilla & 2071.68 $\pm$ 9.83 & 3974.55 $\pm$ 588.24 & 14947.88 $\pm$ 808.89 & 4845.64 $\pm$ 292.69 \\
                     & VIAC    & 2088.18 $\pm$ 208.72 & 4063.18 $\pm$ 1879.02 & 15563.37 $\pm$ 380.11 & 4880.45 $\pm$ 604.25 \\
                     & BEE     & 2114.86 $\pm$ 146.65 & 4682.34 $\pm$ 1487.38 & 15182.46 $\pm$ 536.78 & 4930.32 $\pm$ 794.63 \\
                     & CARE-VI & \textbf{2263.54 $\pm$ 134.62} & \textbf{5585.63 $\pm$ 206.51} & \textbf{16017.68 $\pm$ 738.81} & \textbf{5316.07 $\pm$ 58.88} \\
\midrule
\multirow{4}{*}{TD3} & Vanilla & 3280.65 $\pm$ 284.24 & 4306.09 $\pm$ 560.71 & 11305.54 $\pm$ 857.45 & 4676.71 $\pm$ 246.61 \\
                     & VIAC    & 3460.01 $\pm$ 189.32 & 4651.96 $\pm$ 1535.85 & 13812.05 $\pm$ 573.62 & 5145.74 $\pm$ 207.17 \\
                     & BEE     & 3311.07 $\pm$ 329.92 & 4858.24 $\pm$ 170.63 & 12647.81 $\pm$ 1889.75 & 5024.40 $\pm$ 502.26 \\
                     & CARE-VI & \textbf{3527.22 $\pm$ 91.56} & \textbf{6365.54 $\pm$ 140.49} & \textbf{14278.99 $\pm$ 735.92} & \textbf{5521.98 $\pm$ 405.85} \\
\midrule
\multirow{4}{*}{TD7} & Vanilla & 2190.50 $\pm$ 104.06 & 7450.55 $\pm$ 2218.79 & 17531.65 $\pm$ 240.46 & 5493.29 $\pm$ 1373.93 \\
                     & VIAC    & 2269.46 $\pm$ 96.69 & 8294.86 $\pm$ 1548.94 & 17692.96 $\pm$ 108.05 & 5642.32 $\pm$ 535.16 \\
                     & BEE     & 2481.91 $\pm$ 453.60 & 8487.25 $\pm$ 215.39 & 17489.06 $\pm$ 269.12 & 5554.33 $\pm$ 1114.16 \\
                     & CARE-VI & \textbf{2538.42 $\pm$ 413.09} & \textbf{8511.92 $\pm$ 174.39} & \textbf{17851.35 $\pm$ 114.86} & \textbf{6464.97 $\pm$ 291.01} \\
\bottomrule
\end{tabular}
\end{table*}

\begin{figure}[!h]
    \centering
    \includegraphics[width=0.8\linewidth]{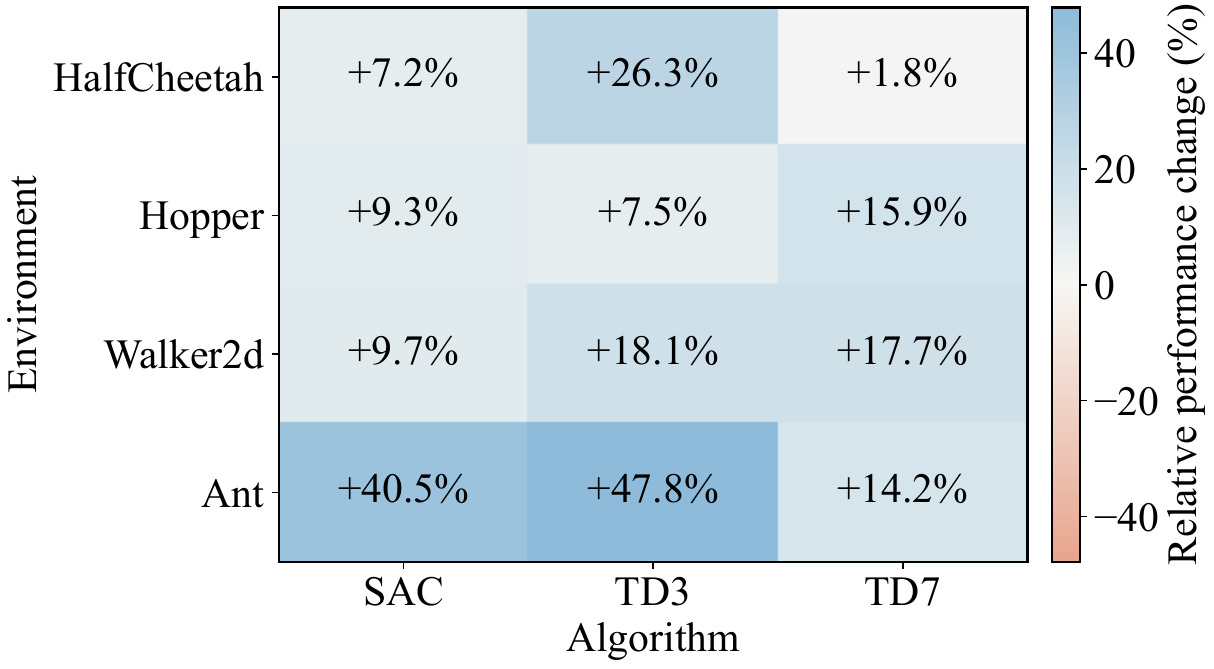}
    \vspace{0cm}
    \caption{Relative return change of CARE-VI over each corresponding Vanilla
    method across the evaluated backbones and MuJoCo tasks.}
    \label{fig:heatmap}
\end{figure}

CARE-VI obtains the highest reported mean return in all twelve tested
backbone-task configurations. The same ordering appears with SAC, TD3, and TD7
despite their different update rules. This cross-backbone pattern provides
method-level evidence for the proposed target construction within the evaluated
continuous-control setting.
Figure~\ref{fig:heatmap} provides the corresponding relative changes over each
Vanilla backbone. All cells are positive, although their magnitudes vary with
the task and backbone.

CARE-VI evaluates multiple candidate actions and a separately parameterized
evaluator, which adds computation to target construction. The additional timing comparison measures when each method
first reaches $95\%$ of the corresponding Vanilla score. CARE-VI reaches this
return threshold earlier in seven of the twelve configurations, showing that
its final-return advantage is accompanied by earlier attainment of the matched
score in a majority of the tested cases.

\subsection{Ablation Analysis}

To investigate the contribution of the target-construction components, we apply
three substitutions while retaining the remaining components and backbone
settings. Without CARS, candidates are ranked by the selector mean and
a fixed $K_{\max}$ prefix is retained. Without SEVA, the retained candidates
are selected by the selector score and the selector mean supplies the reviewed
value. Without the DARE risk gate, the finite stage factor is retained while
the local reliability and evidence-gap attenuation are set to one. These substitutions
leave the remaining target construction unchanged.

\begin{table}[H]
\centering
\caption{Relative return reduction under each CARE-VI component ablation. Each
entry normalizes the backbone-aggregated score by full CARE-VI.}
\label{tab:module-contribution}
\vspace{-2mm}
\small
\begin{tabular}{lcccc}
\toprule
\textbf{Removed component} & \textbf{SAC} & \textbf{TD3} & \textbf{TD7} & \textbf{Mean} \\
\midrule
CARS & $6.8\%$ & $4.8\%$ & $1.4\%$ & $4.4\%$ \\
SEVA & $4.7\%$ & $6.4\%$ & $1.8\%$ & $4.3\%$ \\
DARE risk gate & $8.7\%$ & $5.1\%$ & $0.6\%$ & $4.8\%$ \\
\bottomrule
\end{tabular}
\vspace{-3mm}
\end{table}

Table~\ref{tab:module-contribution} shows that every reported ablation reduces
the backbone-aggregated return, while the largest decrease depends on the
backbone. For each backbone, the table expresses the drop from full CARE-VI in
the same backbone-level aggregate. It therefore
provides a within-backbone summary of the three grouped substitutions. The CARS
and SEVA substitutions affect retained-width control and capped review, while
the DARE substitution isolates evidence-dependent residual attenuation with the
finite schedule held fixed.

\subsection{Mechanism Diagnostics}

To examine the failure modes that motivate the three components, we conduct two
principle-level scalar experiments. The first isolates estimation reuse and
disagreement-aware candidate ranking under localized critic bias. The second
varies the true value landscape and critic noise across actions to compare three
residual enhancement rules.

For the localized bias experiment, we consider
\[
Q^\ast(a)=1-a^2, \qquad a\in[-1,1],
\]
whose maximizing action is $a^\ast=0$. We repeat the experiment for 300
independent runs, each consisting of 120 bootstrapping steps with 64 actions
drawn uniformly from $[-1,1]$ at every step. The two
selector estimates are constructed by adding local overestimation centered at
$a=0.55$, with amplitudes $0.45$ and $0.225$ and a common width of $0.08$.
For the independent evaluator, the overestimation is centered at $a=-0.55$
with amplitude $0.12$ and width $0.10$. Independent zero mean Gaussian noise
with standard deviation $0.05$ is then added to every estimate.

The comparison includes three target construction variants.
\textsc{Coupled-Top1} selects the action with the largest first selector
estimate and reuses that estimate for target valuation, whereas
\textsc{Decoupled-Review} keeps the selected action unchanged and values it
with the independent evaluator. \textsc{Conservative-Decoupled} instead ranks
the actions by the selector mean minus absolute selector disagreement, reviews
the selected action with the evaluator, and caps the reviewed value by the
selector mean. We measure accumulated target bias from $B_0=0$ according to
\[
B_t=0.98B_{t-1}+\bigl(\widehat Q_t(a_t)-Q^\ast(a_t)\bigr).
\]
The quality of candidate selection is recorded separately through the selected
action error $|a_t-a^\ast|=|a_t|$.

\begin{figure*}[!t]
    \centering
    \includegraphics[width=\textwidth]{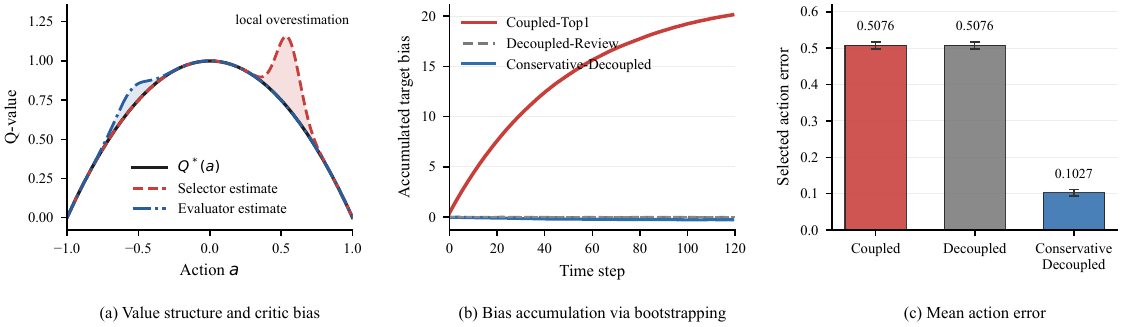}
    \caption{Localized critic bias experiment. From left to right, the panels
    depict the true value and critic estimates, the accumulated target bias
    over 120 bootstrapping steps, and the mean selected action error. Bias
    curves are averaged over 300 independent runs. Error bars denote the
    standard deviation of the mean action error computed within each run.}
    \label{fig:same_source_error}
\end{figure*}

As shown in Fig.~\ref{fig:same_source_error}, \textsc{Coupled-Top1}
accumulates a persistent positive target bias, whereas the variants valued by
the separately generated evaluator keep the bias near zero. Because
\textsc{Coupled-Top1} and \textsc{Decoupled-Review} select the same actions,
their difference isolates the effect of reusing the selection score for target
valuation. Their similar action errors also show that changing the value source
alone does not improve candidate quality. Adding conservative selector
evidence substantially lowers the action error while retaining controlled
target bias. These observations isolate two design requirements used by
CARE-VI, namely separate value review and disagreement-aware candidate ranking.

To further examine how enhancement behaves when local value evidence changes,
we construct a second experiment with three value landscapes. For each
landscape, we generate 5000 trials with 64 candidate actions per trial. A base
action is drawn from a zero mean Gaussian
distribution with standard deviation $0.35$ and clipped to $[-1,1]$. Each
candidate adds an independent zero mean Gaussian perturbation with standard
deviation $0.45$ before the same clipping. The true value functions are
\[
\begin{aligned}
Q^\ast_{\mathrm{flat}}(a)&=1-0.1a^2,\\
Q^\ast_{\mathrm{multi}}(a)&=\max\!\left\{
e^{-30(a-0.2)^2},\,0.9e^{-30(a+0.6)^2}\right\},\\
Q^\ast_{\mathrm{sharp}}(a)&=e^{-50a^2}.
\end{aligned}
\]
Two selector estimates and one evaluator estimate receive independent zero
mean Gaussian noise whose standard deviation varies with the action as
$\sigma(a)=0.04+0.25|a|$.

\begin{figure*}[!t]
    \centering
    \includegraphics[width=\textwidth]{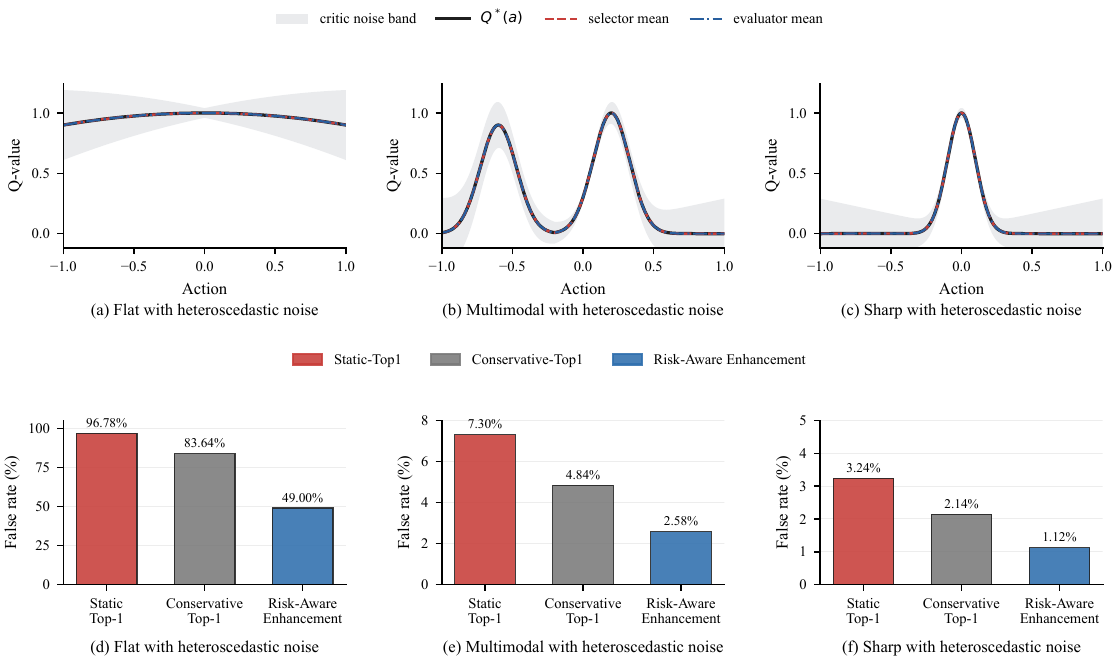}
    \caption{False enhancement under critic noise that varies across actions. The
    top row presents the flat, multimodal, and sharp value landscapes together
    with their noise bands. The bottom row compares the false enhancement rates
    of the three rules over 5000 trials.}
    \label{fig:false_improvement}
\end{figure*}

Three enhancement rules are evaluated under the same sampled actions.
\textsc{Static-Top1} chooses the largest first selector estimate and applies an
enhancement coefficient of one. \textsc{Conservative-Top1} replaces this
ranking with the selector mean minus absolute disagreement but retains the
same fixed coefficient. The rule labeled \textsc{Risk-Aware Enhancement} in
Fig.~\ref{fig:false_improvement} combines the three corresponding principles in
this scalar setting. It retains an adaptive prefix of at most eight actions,
fuses normalized selector and evaluator signals with equal weights, and caps
the reviewed value by the selector mean. Its evidence-dependent coefficient is
\[
\lambda=\max\!\left\{0.05,\exp(-3u-1.5d_{\mathrm{gap}})\right\},
\]
where $u$ denotes selector disagreement at the selected action and
$d_{\mathrm{gap}}$ is the absolute gap between its normalized selector and
evaluator signals. We
define a false enhancement as a trial in which the estimated gain is positive,
$\widehat G>0$, but the corresponding true gain is negative, $G<0$. The false
enhancement rate is obtained by averaging this event over the 5000 trials.

Across the three landscapes, conservative ranking reduces false enhancement
relative to \textsc{Static-Top1}, and \textsc{Risk-Aware Enhancement} produces
the lowest rate. The difference is most informative on the flat landscape,
where true value gaps are weak relative to critic noise. In this regime, the
adaptive rule assigns less weight to ambiguous evidence instead of converting
a small estimated advantage directly into a full target correction. On the
multimodal and sharp landscapes, \textsc{Conservative-Top1} obtains a larger
mean true gain but also accepts false enhancements more often. The comparison
isolates a distinction between choosing a candidate and determining the size of
its target correction. This distinction motivates the evidence-dependent
residual scaling used by DARE.

\subsection{Sensitivity Analysis}

\begin{figure}[!t]
    \centering
    \includegraphics[width=0.95\linewidth]{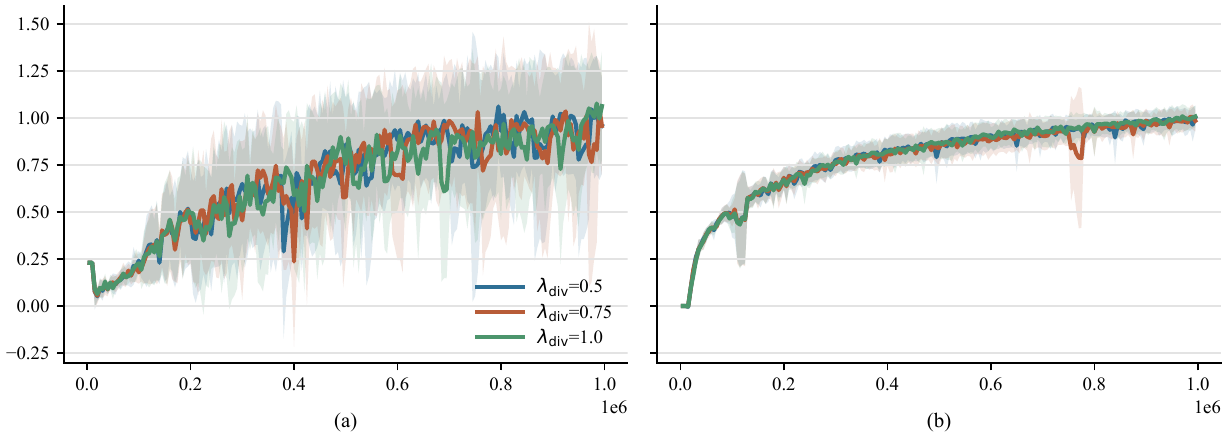}
    \caption{Normalized learning curves under different CARS disagreement
    penalties $\lambda_{\mathrm{div}}$ on Ant-v4 in the left panel and
    HalfCheetah-v4 in the right panel. The horizontal axis shows environment steps and the vertical axis shows normalized return.}
    \label{fig:sensitivity_rank_penalty}
\end{figure}

\begin{figure}[!t]
    \centering
    \includegraphics[width=0.95\linewidth]{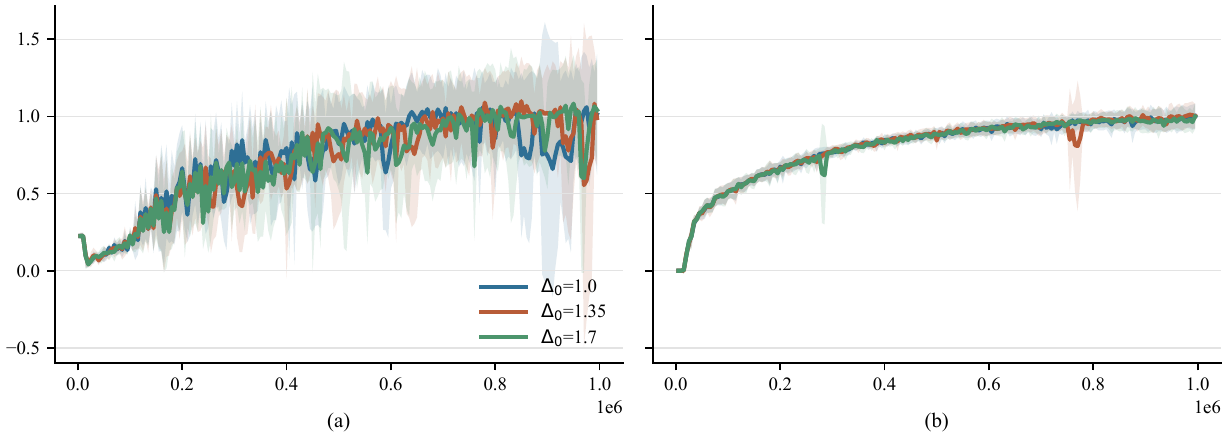}
    \caption{Effect of the DARE gap tolerance $\Delta_0$ on normalized learning
    curves for Ant-v4 in the left panel and HalfCheetah-v4 in the right panel. The horizontal axis shows environment steps and the vertical axis shows normalized return.}
    \label{fig:sensitivity_gap_target}
\end{figure}

We vary the CARS disagreement penalty $\lambda_{\mathrm{div}}$ and the DARE
gap tolerance $\Delta_0$ separately while keeping all other settings fixed.
Each sweep uses $1\times10^6$ environment steps.
Figures~\ref{fig:sensitivity_rank_penalty} and
\ref{fig:sensitivity_gap_target} compare normalized learning trajectories
under the tested values on Ant-v4 and HalfCheetah-v4.

The trajectories remain close through most of training, and no setting is
uniformly best on both tasks. Within the tested ranges, the gains are therefore
not confined to one narrow value of either representative control. The two
sweeps also agree with their intended roles. The parameter
$\lambda_{\mathrm{div}}$ changes how ranking responds to selector
disagreement, whereas $\Delta_0$ changes when the selector-evaluator gap begins
to attenuate the residual.

\subsection{DARE Schedule Analysis}

\begin{figure}[!t]
    \centering

    \includegraphics[width=0.98\linewidth]{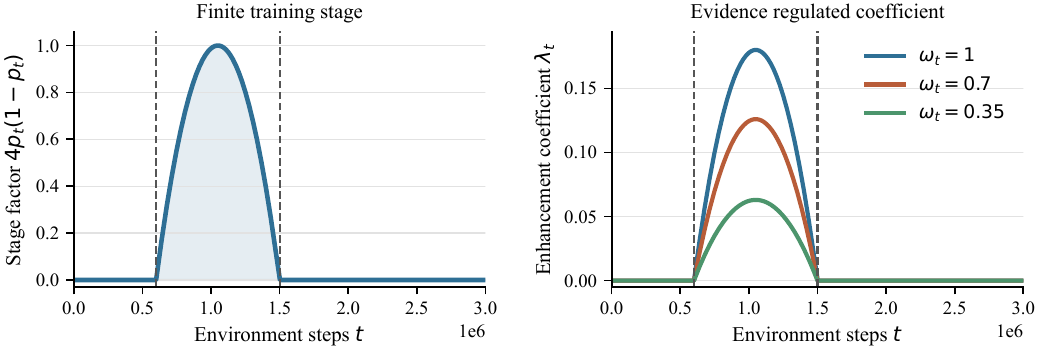}
    \caption{Finite DARE schedule with the quadratic stage factor
    $4p_t(1-p_t)$ in the left
    panel and enhancement coefficients under representative risk gates in the
    right panel. Dashed lines mark the endpoints of the active interval.}
    \label{fig:dare_dynamics}
\end{figure}

Training stage and local evidence serve different roles in DARE.
Figure~\ref{fig:dare_dynamics} visualizes the stage factor in
\eqref{eq:dare_training_stage} and the enhancement coefficient in
\eqref{eq:dare-enhancement-coef} under the experimental schedule.

The left panel shows that the stage factor is confined to the active interval,
reaches its maximum inside the interval, and vanishes at both endpoints. In the
right panel, a lower risk gate reduces the coefficient without extending this
support. The two factors therefore control different axes of the correction.
The stage factor determines when the residual can act, while the risk gate
determines its magnitude from local evidence. This analytic property supplies
the finite support used by the fixed-policy recovery result.

\section{Conclusion}

This article develops evidence-regulated direct value improvement for
bootstrapped target construction. CARS limits candidate commitment with an
ordered prefix and a disagreement-scaled uncertainty radius. SEVA combines
selector-led ordering with auxiliary evaluator evidence and caps the reviewed
value by the selector reference. DARE then regulates the magnitude and duration
of the resulting residual from candidate reliability, the selector-evaluator
gap, and training stage. CARE-VI integrates these components while retaining the
critic-regression and actor-update interfaces of the underlying actor-critic
backbone. The analysis bounds CARS boundary error, SEVA selected-value
overestimation, and the one-sided deviation of the DARE residual displacement from its
population counterpart. A finite-horizon evaluation bound further shows how
the finite CARE-VI perturbation and fitting residual affect recovery to the base
fixed-policy solution. In the twelve combinations formed by SAC, TD3, TD7, and
four MuJoCo tasks, CARE-VI records the highest reported mean return in every
tested configuration.
The component ablations and controlled studies connect this cross-backbone
pattern to the grouped target components and to the failure modes that motivate
capped review and evidence-dependent residual scaling. These results establish
target-specific evidence regulation as a practical way to improve direct value
construction within the evaluated continuous-control setting.

\bibliographystyle{IEEEtran}
\bibliography{cite}

\end{document}